\documentclass[review]{elsarticle}

\usepackage{float}
\usepackage{lineno,hyperref}
\usepackage[title,titletoc,toc]{appendix}
\usepackage{makecell}
\modulolinenumbers[5]

\usepackage{xcolor}   % for comments in the text
\usepackage{booktabs}
\makeatletter
\def\ps@pprintTitle{%
  \let\@oddhead\@empty
  \let\@evenhead\@empty
  \let\@oddfoot\@empty
  \let\@evenfoot\@oddfoot
}
\makeatother

\biboptions{authoryear}

\begin{document}

\begin{frontmatter}

\title{Transfer Learning with Ensembles of Deep Neural Networks for Skin Cancer Detection in Imbalanced Data Sets\tnoteref{mytitlenote}}
%%\tnotetext[mytitlenote]{Fully documented templates are available in the elsarticle package on \href{http://www.ctan.org/tex-archive/macros/latex/contrib/elsarticle}{CTAN}.}
%%\author{Elsevier\fnref{myfootnote}}
%%\address{Radarweg 29, Amsterdam}
%%\fntext[myfootnote]{Since 1880.}

\author[mymainaddress]{Aqsa Saeed Qureshi\corref{mycorrespondingauthor}}
\ead{aqsa.qureshi@helsinki.fi}
\author[mymainaddress]{Teemu Roos}
%%\cortext[mycorrespondingauthor]{Aqsa Saeed Qureshi}
%%\ead{support@elsevier.com}
\address[mymainaddress]{Department of Computer Science, University of Helsinki, Finland.}
\begin{abstract}
%Skin cancer is among one of the most common cancers across the world. 
Early diagnosis plays a key role in prevention and treatment of skin cancer. 
%All the clinical diagnosis methods take time therefore, Machine Learning (ML) techniques are widely used to automatically detects skin cancer and its type. 
Several machine learning techniques for accurate detection of skin cancer from medical images have been reported. 
Many of these techniques are based on pre-trained convolutional neural networks (CNNs), which enable training the models based on limited amounts of training data. However, the classification accuracy of these models still tends to be severely limited by the scarcity of representative images from malignant tumours. We propose a novel ensemble-based CNN architecture where multiple CNN models, some of which are pre-trained and some are trained only on the data at hand, along with auxiliary data in the form of metadata associated with the input images, are combined using a meta-learner. The proposed approach improves the model's ability to handle limited and imbalanced data. We demonstrate the benefits of the proposed technique using a dataset with 33126 dermoscopic images from 2056 patients. 
%After the training, all the features extracted from base-learners along with the meta-data is provided as input to Support Vector Machine (SVM) which finally classify the input image. 
We evaluate the performance of the proposed technique in terms of the F1-measure, area under the ROC curve (AUC-ROC), and area under the PR-curve (AUC-PR), and compare it with that of  seven different benchmark methods, including two recent CNN-based techniques. The proposed technique compares favourably in terms of all the evaluation metrics.
%, even though the differences between the top performing methods fit within statistical margin of error.
\end{abstract}
\begin{keyword} 
\texttt Skin cancer; Deep learning; Transfer learning; Ensemble methods
\end{keyword}
\end{frontmatter}
\section{Introduction}
Skin cancer is caused by mutations within the DNA of skin cells, which causes their abnormal multiplication  \citep{armstrong1995skin,simoes2015skin}. 
%Exposure of skin to sunlight is the most common cause of skin cancer but it can also develop on the unexposed areas of skin \citep{saladi2005causes}. Besides UV radiations other factors also cause skin cancer; like people having fair complexion, abnormal history of having moles, family history of having moles, weak immunity, exposure to certain substances are at higher risk to develop skin cancer \citep{tl2002epidemiology,5}. Change in size of mole, irregular shaped lesion, lesion with itching or pain, brownish spots on skin are some of the common signs of skin cancer.
In the early development of skin cancer, lesions appear on the the outer layer of the skin, the epidermis.  Not all lesions are caused by malignant tumours, and a diagnosis classifying the lesion as either malignant (cancerous) or benign (non-cancerous) is often reached based on preliminary visual inspection followed by a biopsy. Early detection and classification of lesions is important because early diagnosis of skin cancer significantly improves the prognosis~\citep{7}.

%Unfortunately, the whole procedure is time consuming and patient may develop the later stage of cancer. 
The visual inspection of potentially malignant lesions carried out using an optical dermatoscope is a challenging task and requires a specialist dermatologist. For instance, according to~\cite{6}, in the case of melanoma, a particularly aggressive type of skin cancer, only about 60--90~\% of malignant tumours are identified based on visual inspection, and accuracy varies markedly depending on the experience of the dermatologist. As skillful dermatologists are not available globally and for all ethnic and socioeconomic groups, the situation causes notable global health inequalities~\citep{buster2012}. 

Due to the aforementioned reasons, machine learning techniques are widely studied in the literature. Machine learning has potential to aid automatic detection of skin cancer from dermoscopic images, 
%and the type of cancer, 
thus enabling early diagnosis and treatment. %\citep{8, ballerini2013color, thomas2021interpretable,9, 10, 11, 12, 13, 14, 15, 16, 17, 18, lei2020skin}. %For detection of skin cancer, different data pre-processing techniques have been applied to the input images, generated with the help of dermatoscope. 
\cite{8} compared the performance of K-nearest neighbor (KNN), random forest (RF), and support vector machine (SVM) classifiers on data extracted from segmented regions of demoscopic images. 
%Results showed that SVM performed well in comparison to other classifiers for skin cancer classification.
Similarly, \cite{ballerini2013color} used a KNN-based hierarchical approach for classifying five different types of skin lesions. \cite{thomas2021interpretable} used deep learning based methods for classification and segmentation of skin cancer. 
%For the early diagnosis of skin cancer, 
\cite{9} proposed a 
%diagnostic 
%system that first enhanced the input images and only those regions are left in the images which hold cancerous cells. After that, information from the enhanced regions is then pass to 
technique based on a
Multi-Layer Perceptron (MLP) and other neural network models. A recent review by \cite{chan2020machine} summarizing many of these studies concluded that while many authors reported better sensitivity and specificity than dermatologists, ``further validation in prospective clinical trials in more real-world settings is necessary before claiming superiority of algorithm performance over dermatologists.''
%classifier and auto-associated Neural Network (NN). Results showed that MLP performed better in terms of accuracy in comparison to auto associated NN. 
%%\textbf{Can we say something generic about the achieved accuracy, like "In these studies, the accuracy of machine learning based techniques was XYZ-ZYX", or something?}

What all the aforementioned methods have in common is that they require large amounts of training data in the form of dermoscopic images together with labels indicating the correct diagnosis. Several authors have proposed approaches to reduce the amount of training data required to reach satisfactory classification accuraracy.
\cite{10} describe a method based on \emph{transfer learning}, which is a way to exploit available training data collected for a different classification task than the one at hand. Hosny’s technique is based on a pre-trained AlexNet network (a specific deep learning architecture proposed by~\cite{alexnet2012}) that was originally trained to classify images on a commonly used ImageNet dataset, and then adapted to perform skin cancer classification by transfer learning. 
Similarly, \cite{11} used a pre-trained AlexNet  combined with a SVM classifier. \cite{13} utilized a ResNet network, another commonly used deep learning architecture by~\cite{22}. \cite{14} use a combination of two deep learning models for the segmentation and classification of skin lesions. %Convolutional Neural Network (CNN) is used for extracting features from dataset. 
%The dataset used in Li’s technique was highly imbalanced therefore data augmentation was performed prior to segmentation and classification.
\cite{15} suggested a transfer learning based technique in which hyperspectral data is used for the detection of melanoma using a pre-trained GoogleNet model~\citep{17}. The same pre-trained model is used by \cite{16}. 
%In \cite{18} technique for skin cancer detection, a three-stage framework is designed. In first step contrast enhancement is performed on the input images, whereas in second step color CNN approach is used to extract the boundary of the skin lesion. In third step 
\cite{19} use a pre-trained Inception V3 model~\citep{inception}. 

Another commonly used technique to improve classification accuracy when limited training data is available is \emph{ensemble learning}, see \citep{ensemble1,ensemble2}. The idea is to combine the output of multiple classifiers, called \emph{base-learners}, by using their outputs as the input to another model that is called a \emph{meta-learner}, to obtain a consensus classification. Ensemble learning tends to reduce variability and improve the classification accuracy. \cite{12} proposed an ensemble based hybrid technique
%. In this technique first pre-processing and data augmentation is performed on input data. After that, three set of features are extracted from the augmented data. First and second sets of feature spaces are extracted from fully connected layers of
involving pre-trained AlexNet, VGG16 \citep{21}, and ResNet models as base-learners. Output obtained from these models is combined using SVM and logistic regression classifiers as meta-learners.
%Whereas third set of feature space is extracted from last fully connected and convolutional layers of pre-trained ResNet18 \citep{22}. 
%After that three different SVM’s are trained on all the different extracted feature spaces. In the end all the predicted values from SVMs are than provided to simple logistic regression unit which finally performs classification. 
%In another transfer learning based approach, \cite{19} performed fine-tuning of pre-trained InceptionV3 network on skin cancer classification dataset. During training phase all images are resized to $229\times229$ dimensions to made them compatible with the input dimensions of InceptionV3 architecture.

In addition to the shortage of large quantities of labeled training data, many clinical datasets have severe \emph{class imbalance}: the proportion of positive cases tends to be significantly lower than that of the negative cases, see~\citep{he-garcia-2009}. This reduces the amount of informative data points and lowers the accuracy of many machine learning techniques, and may create a bias that leads to an unacceptably high number of false negatives when the model is deployed in real-world clinical applications. To deal with the class imbalance issue, most of the previously reported techniques use \emph{data augmentation}, i.e., oversampling training data from the minority (positive) class and/or undersampling the majority class. This tends to lead to increased computational complexity as amount of training data is in some cases increased many-fold, and risks losing informative data points due to undersampling.

%Also, most of the techniques fine-tuned the pre-trained networks for improving the classification accuracy. Disadvantage of using the pre-trained network is that these networks are trained on random set of images, which may lead to negative learning in predicting cancerous images in target domain. Also, all the previously reported skin cancer classification techniques used the images, there is no auxiliary patient information used along with the images. 

In this paper, we propose a new technique for skin cancer classification from dermoscopic images based on transfer learning and ensembles of deep neural networks. The motivation behind the proposed technique is to maximize the diversity of the base-classifiers at various levels during the training of the ensemble to improve the overall accuracy.  In our proposed ensembled-based technique, a number of CNN base-learners are trained on input images scaled to different sizes between $32\times 32$ and $256\times 256$ pixels. During training, two out of six base-learners are pre-trained CNNs trained on another skin cancer dataset that is not part of primary dataset. In the second step, all the predictions from base-learners along with the meta-data, including, e.g., the age and gender of the subject, provided with the input images is provided to an SVM meta-learner to obtain the  final classification. By virtue of  training the base-classifiers on different input images of different sizes, the model is able to focus on features in multiple scales at the same time. The use of meta-data further diversifies the information which improves the classification accuracy. 

We evaluate the performance of the proposed technique on data from the International Skin Imaging Collaboration (ISIC) 2020 Challenge, which is highly imbalanced containing less than 2~\% malignant samples \citep{data}. Our experiments demonstrate that $(i)$ ensemble learning significantly improves the accuracy even though the accuracy of each of the base-learners is relatively low; $(ii)$ transfer learning and the use of meta-data have only a minor effect on the overall accuracy; $(iii)$ overall, the proposed method compares favourably against to all of the other methods in the experiments, even though the differences between the top performing methods fit with statistical margin of error.

\section{The Proposed Method}

\begin{figure}[H]
    \centering
    \includegraphics[width=1\textwidth]{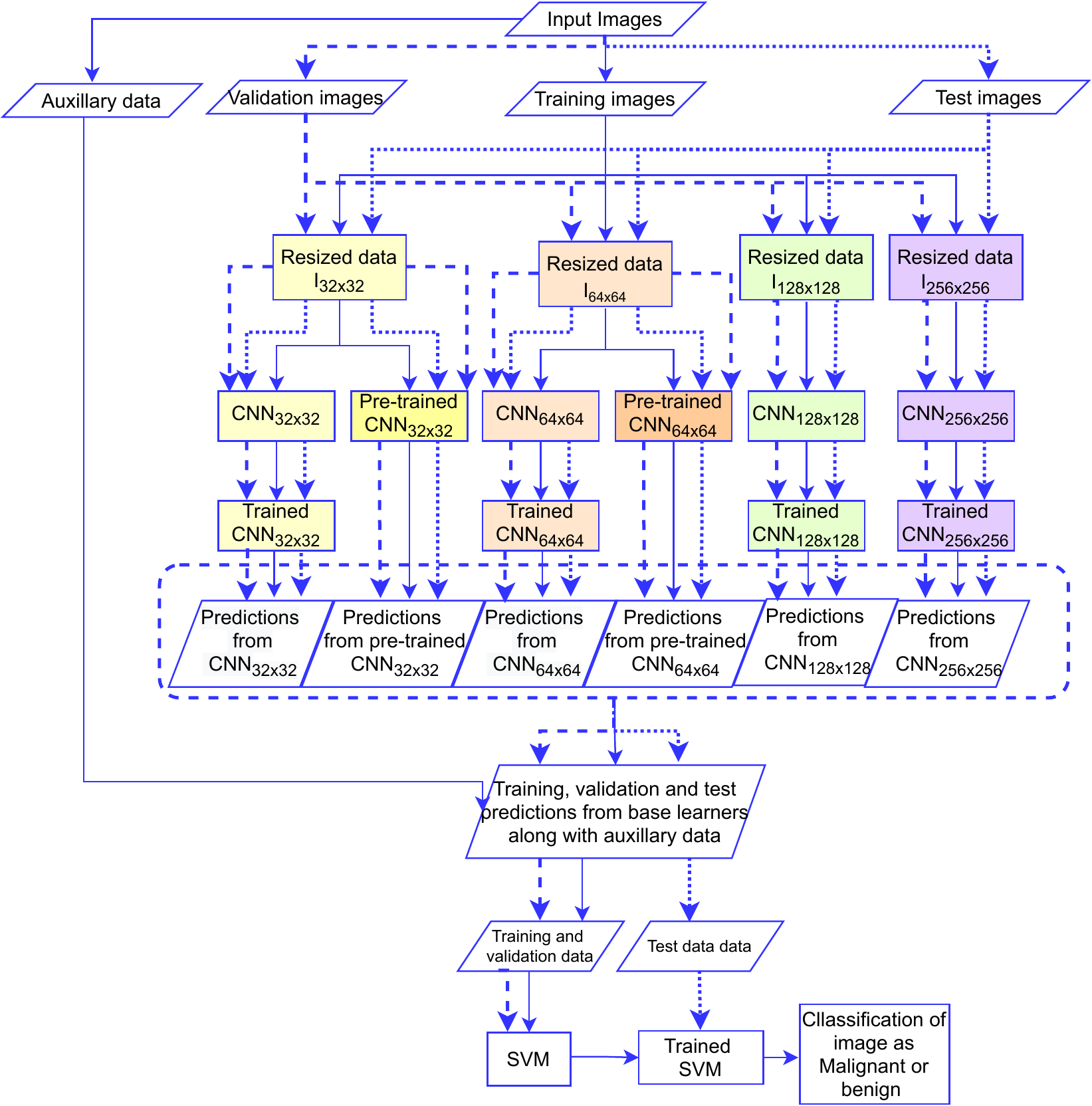}
    \caption{Block diagram of the proposed technique}
    \label{block}
\end{figure}

The proposed technique is an ensemble-based technique in which CNNs are used as base learners. The base learners are either pre-trained on balanced dataset collected from ISIC archive or on the the ISIC 2020 dataset. Predictions from all the base learners along with the auxiliary data contained in the metadata associated to the images are used as input to an SVM classifier, which finally classifies each image in dataset as positive (malignant) or negative (benign). Figure \ref{block} shows the flowchart of the proposed technique. 

\subsection{Architecture}

In the proposed technique, six base-learners are used. Each of the base learners operates on input data of different dimensions. During training four base learners, $\mathrm{CNN}_{32\times32}$, $\mathrm{CNN}_{64\times64}$, $\mathrm{CNN}_{128\times128}$, and $\mathrm{CNN}_{256\times256}$, are trained from random initial parameters on $32\times 32$, $64\times 64$, $128\times 128$, and $256\times 256$ input images respectively. Another two base learners are trained on  malignant and benign skin cancer images of sizes $32\times 32$ and $64 \times 64$ respectively, which are not part of ISIC 2020 dataset. After training of all six base-learners, predictions from all of them, along with the metadata is then fed into an SVM classifier that functions as the meta classifier. The SVM is trained on the training data and used to finally classify each of the test images as malignant or benign. For both the base and the meta classifiers, the validation data is used to adjust hyperparameters as explained in more detail below. 

\subsection{Transfer learning during training of the base-learners}

Transfer learning is used to transfer the knowledge extracted from one type of machine learning problem to another~\citep{TL,survey_TL}. The domain from where information is extracted is known as the source domain, and the domain where extracted information is applied is called the target domain. The benefit of transfer learning is that it not only saves time that is needed to train the network from scratch but also aid in improving the performance in the target domain. 
%There are diverse types of transfer learning approaches and each type depends on the way knowledge is transferred from source domain to target domain \citep{.

In the proposed technique, idea of transfer learning is exploited during the training phase of base-learners. 
%In machine learning to deal with data imbalance mostly down sampling is used, in which majority class is down sample to make balance data set for the training of classifier. The disadvantage of down sampling is that it may lead to loss of some valuable information from data. Another way to balance the data set is to augment the samples which are present in minority, but disadvantage of data augmentation is that it can only extracts limited information, and the information is not as much precise as real data have. Therefore, in this proposed technique instead of losing any information from data all the base-learners are trained on real data samples. 
%The ISIC 2020 dataset used in the proposed technique is highly imbalanced therefore pre-trained 
We pre-train some of the CNN base-learners on a balanced dataset available on Kaggle\footnote{\url{https://www.kaggle.com/fanconic/skin-cancer-malignant-vs-benign}} collected from the ISIC archive is used. This archive dataset was constructed in 2019, so none of the ISIC 2020 data are included in it. The rest of the base-learners are trained on the ISIC 2020 dataset that comprises the target domain. The introduction of the CNNs pre-trained on balanced data provides a diverse set of predictions, complementing the information coming from the base-learners trained on the ISIC 2020 data. Moreover, since the pre-trained base-learners need to be trained only once instead of re-training them every time we repeat the experiment on random subsets of the ISIC 2020 data (see Sec.~\ref{sec:division-of-data} below), the pre-training saves time.

%All the base-learners are trained differently, and all the feature extracted from pre-trained base learners and base-learners trained from scratch on ISIC 2020 data set provide the diverse feature space to meta-classifier.

%\subsection{SVM as a meta-classifier}

%In the proposed technique SVM is used as a meta-learner. After the training of the base learners, all the predicted values along with the meta data is feed as input to the SVM. After that SVM predicts the label of input data as malignant or benign. As SVM is taking diverse information from different base-learners along with some side information contained in the metadata. %, that is why it generalizes well.

\section{Data and Evaluation Metrics}

We use the ISIC 2020 Challenge dataset to train and test the proposed method along with a number of benchmark methods and evaluated their performance with commonly used metrics designed for imbalanced data.

\subsection{Dataset and pre-processing}

The dataset used in the proposed technique contain 33126 dermoscopic images collected from 2056 patients \citep{data}. All the images are labelled using histopathology and expert opinion either as benign or malignant skin lesions. %The dataset is generated by ISIC and images are taken from the below mentioned sources\citep{data}.
%\begin{itemize}
%    \item Hospital clinic Barcelona, Medical University of Vienna.
%    \item University of Athens Medical School.
%    \item Melanoma Institute Australia.
%\end{itemize}
The ISIC 2020 dataset also contains another 10982 test data images without the actual labels, but since we are studying the supervised classification task, we use the labeled data only. All the images are in JPG format of varying dimensions and shape. We use different input dimensions for different base-learners, so we scale the input images to sizes $32\times32$, $64\times64$, $128\times128,$ and $256\times256$ pixels.\footnote{Some of the raw images are non-square shaped (circular or rectangular), in which case we reshape them to make the bounding box square and of the desired size.}

\begin{figure}[H]
    \centering
    \includegraphics[width=1\textwidth]{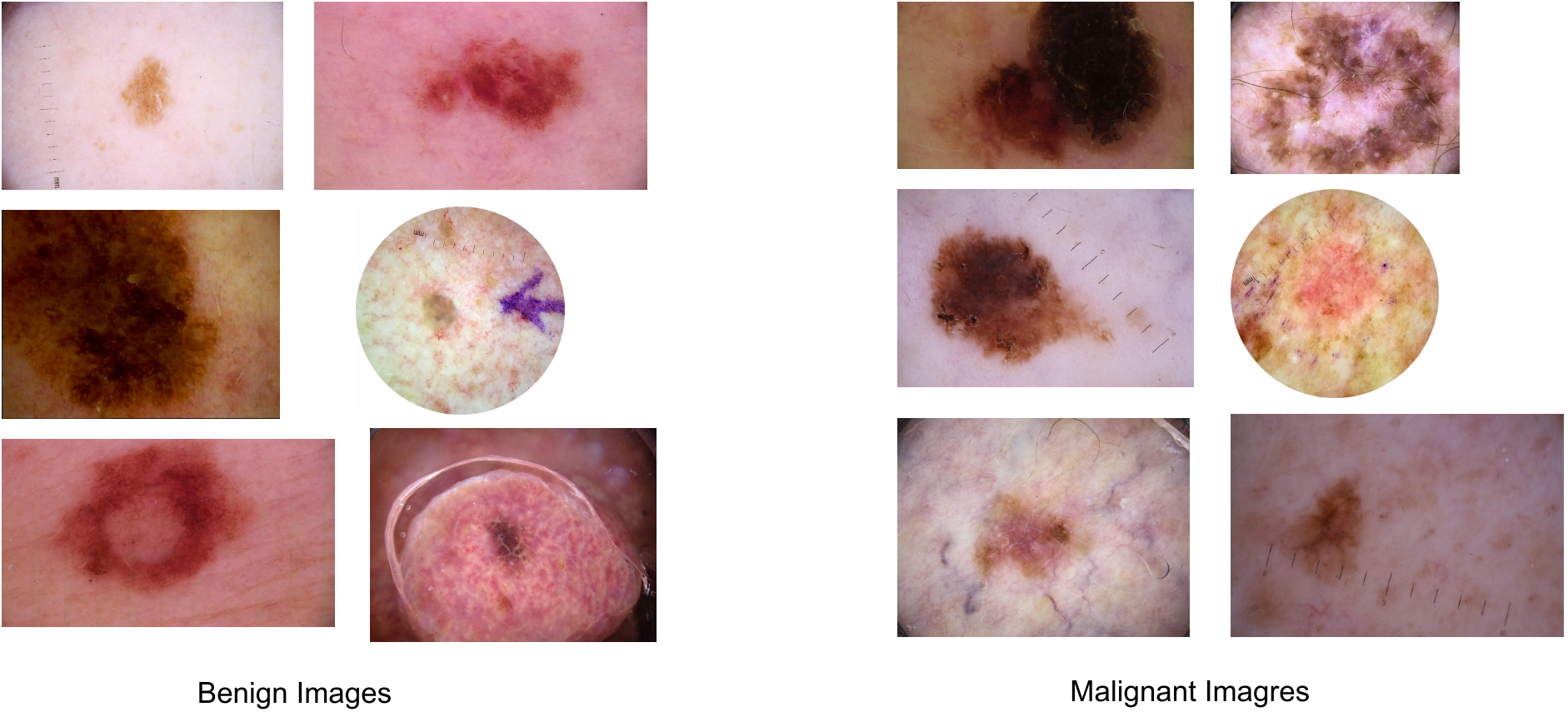}
    \caption{Benign vs malignant images}
    \label{cancer-noncancer}
\end{figure}

Figure~\ref{cancer-noncancer} contains example images present in the dataset. Table~\ref{tab:metatable} shows the features in the metadata. Categorical features were encoded as integers in order to reduce the number of parameters in the meta-learner. All the missing values in the metadata are replaced by the average value of the feature in question.

The ISIC 2020 data set is highly imbalanced because out of the total 33126 images (2056 patients), only 584 images (corresponding to 428 distinct patients) are malignant. The division of the data into training, validation, and test sets is described below in Sec~\ref{sec:division-of-data}.

\begin{table}[H]
\caption{Metadata}\label{tab:metatable} % title of Table
\centering
 \begin{tabular}{rl} 
 \toprule
 Variable & Summary\\ [0.5ex]\midrule
 %\emph{patient ID} & numerical index \\ 
 \emph{age} & min: $10$, max: $90$, median: $50$\\
 \emph{sex} & female: 48.2 \%, male:  51.6\%, other/unknown: 0.2 \%\\
 \emph{general anatomic site} & \makecell{[head/neck, upper extremity, lower extremity,torso,\\ palms/soles, oral/genital]} \\\bottomrule
\end{tabular}
\label{Table1.} % is used to refer this table in the text
\end{table}

%\subsection{Data preprocessing}

%In the proposed technique some of the pre-processing was performed on the images and on meta data provided in ISIC 2020 data set. 

\subsection{Division of the data and hyperparameter tuning}
\label{sec:division-of-data}

\begin{figure}[H]
    \centering
    \includegraphics[width=.5\textwidth]{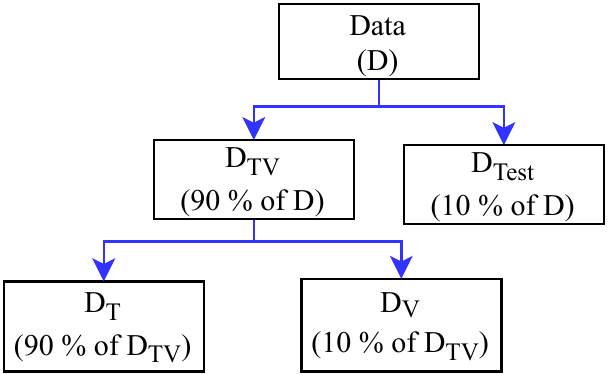}
    \caption{Division of the data in training, $D_\mathrm{T}$, validation $D_\mathrm{V}$, and test $D_\mathrm{Test}$ sets.}
    \label{division}
\end{figure}

We use the validation set method to divide the data in three parts. As illustrated in Fig.~\ref{division}, 10~\% of the total data $D$ is kept as test data, $D_\mathrm{Test}$, which is not used in the training process. The other $90~\%$ is further split by using 90~\% of it as training data, $D_\mathrm{T}$, and the final part as validation data, $D_\mathrm{V}$ which is used to tune the hyperparameters of each of the used methods. Since the ISIC 2020 dataset contains multiple images for the same patient, we require that input images from a given individual appear only in one part of the data ($D_\mathrm{T}$, $D_\mathrm{V}$, or $D_\mathrm{Test}$)\footnote{We use the \texttt{GroupKFold} method in \texttt{scikit-learn} package~\citep{scikit-learn} to do the splitting.}. 
The validation data is used to adjust the hyperparameters in each of the methods in the experiments by maximizing the F1-score (see Sec.~\ref{sec:evalaution} below). Hyperparameter tuning was done manually starting from the settings proposed by the original authors (when available) in case of the compared methods, and adjusting them until no further improvement was observed.
Likewise, the neural network architectures of the CNN base-learners used in the proposed method were selected based on the same procedure as the other hyperparameters. Tables \ref{CNN1}--\ref{CNN6} in the Appendix show the details of the architectures of the CNN base-learners as well as the hyperparameters of the SVM meta-learner.

We used the training and validation data from one random train-validation-test split to tune all hyperparameters. We then used the obtained settings in 10 new random repetitions with independent splits to evaluate the classification performance in order to avoid bias caused by overfitting.

%\subsection{Parameters of deep CNN used as base-learner in the proposed technique}

%\subsection{Parameters of SVM used as meta classifier in the proposed technique}
%In the proposed technique all the predictions from base-learners along with the metadata is provided as input to the SVM, which finally classifies the data as malignant or benign. Table \ref{svm-para} shows the parameters of SVM during the training phase.

\subsection{Evaluation metrics}
\label{sec:evalaution}

As is customary in clinical applications with imbalanced datasets, we use the F1-measure, the area under the ROC curve (AUC-ROC), and the area under the precision--recall curve (AUC-PR) as evaluation metrics; see, e.g., ~\citep{he-garcia-2009}. The F1-measure is the harmonic mean of precision and recall (see definitions below), which is intended to balance the risk of false positives and false negatives. To evaluate the optimal F1-value, we set in each case the classification threshold to the value that maximizes the F1-measure.  

AUC-ROC and AUC-PR both characterize the behavior of the classifier over all possible values of the classification threshold.  AUC-ROC is the area under the curve between the true positive rate (TPR) and the false positive rate (FPR) at different values of the classification threshold, whereas AUC-PR is the area under the precision--recall curve. While both measures are commonly used in clinical applications, according to~\cite{davis2006relationship} and \cite{saito-remsmeier-2015}, AUC-PR is the preferred metric in cases with imbalanced data where false negatives are of particular concern.

The used metrics are defined as follows:
\begin{equation}\label{eq2}
\mbox{Precision}=\frac{T_p}{T_p+F_p}
\end{equation}
\begin{equation}\label{eq3}
\mbox{Recall}=\mathrm{TPR}=\frac{T_p}{P}
\end{equation}
\begin{equation}\label{eq4}
\mathrm{FPR}=\frac{F_p}{N}
\end{equation}
\begin{equation}\label{eq5}
\mbox{F1-measure}=2 \times \frac{\mbox{Precision}\times \mbox{Recall}}{\mbox{Precision}+\mbox{Recall}},
\end{equation}
where $T_P$ is the number of true positives (positive samples that are correctly classified by the classifier), $F_P$ is number of  false positives (negative samples incorrectly classified as positive), and $P$ and $N$ are the the total number positive and negative samples, respectively.

\section{Experimental Results}

%The proposed technique is an ensemble learning based approach, in which training of the base-learners is performed and after that meta-classifier predicts the final class label based on the information collected from the meta data and base-learners. To validate the performance of the proposed technique, holdout cross validation is used. To check the performance of simple base-line classifiers on ISIC 2020 data same training, validation, and test split is used as in the proposed technique. 

All the computations were done on the Puhti supercomputer Atlos Bull\-Sequana X400 cluster comprised of Intel CPUs. 
%All the CPUs relate to 700 nodes having range of storage and memory options. 
For implementation of the deep learning models we use the Keras version 2.2.4 and TensorFlow version 1.14.0. The other machine learning methods and preprocessing methods were implemented in Python 3.0 and \texttt{scikit-learn} version 0.15.2. All the source code needed to replicate the experiments will be released together with the published version of this paper.

\subsection{Main experimental results}

\begin{figure}[tbp!]
    \centering
    \includegraphics[width=1\textwidth]{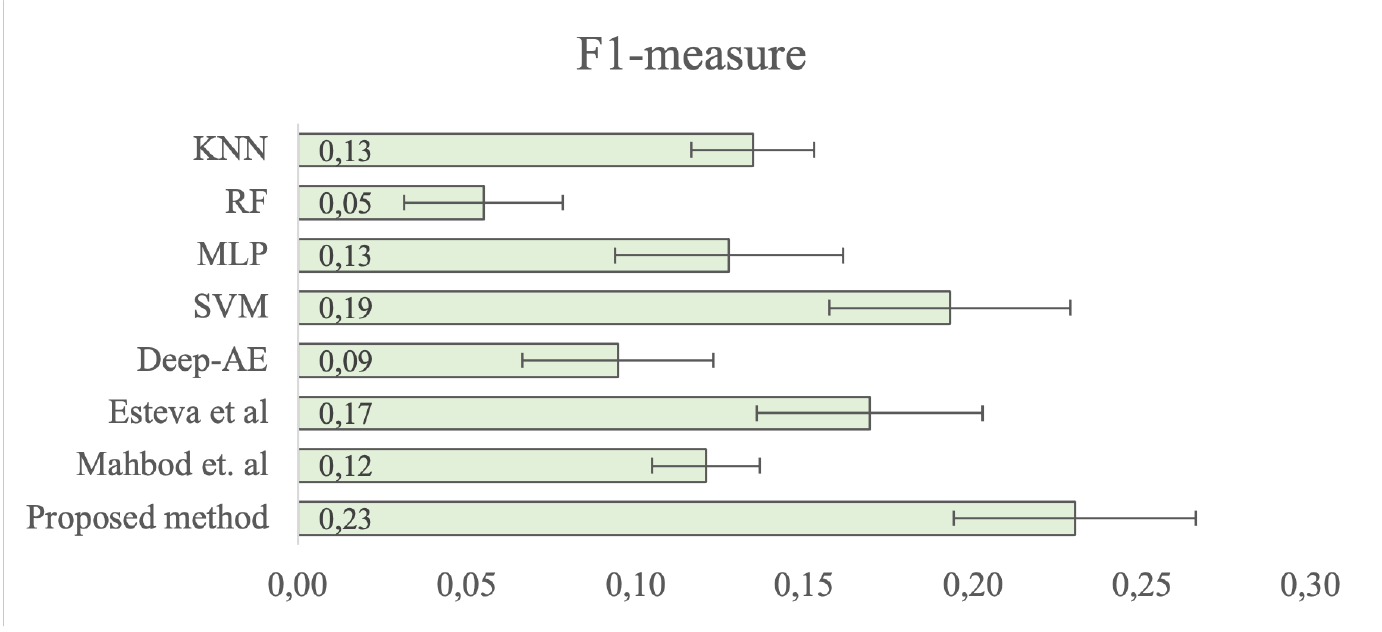}
%    \caption{Comparison of F1-measure over ten independent runs}
%\end{figure}
%\begin{figure}[H]
%    \centering
    \includegraphics[width=1\textwidth]{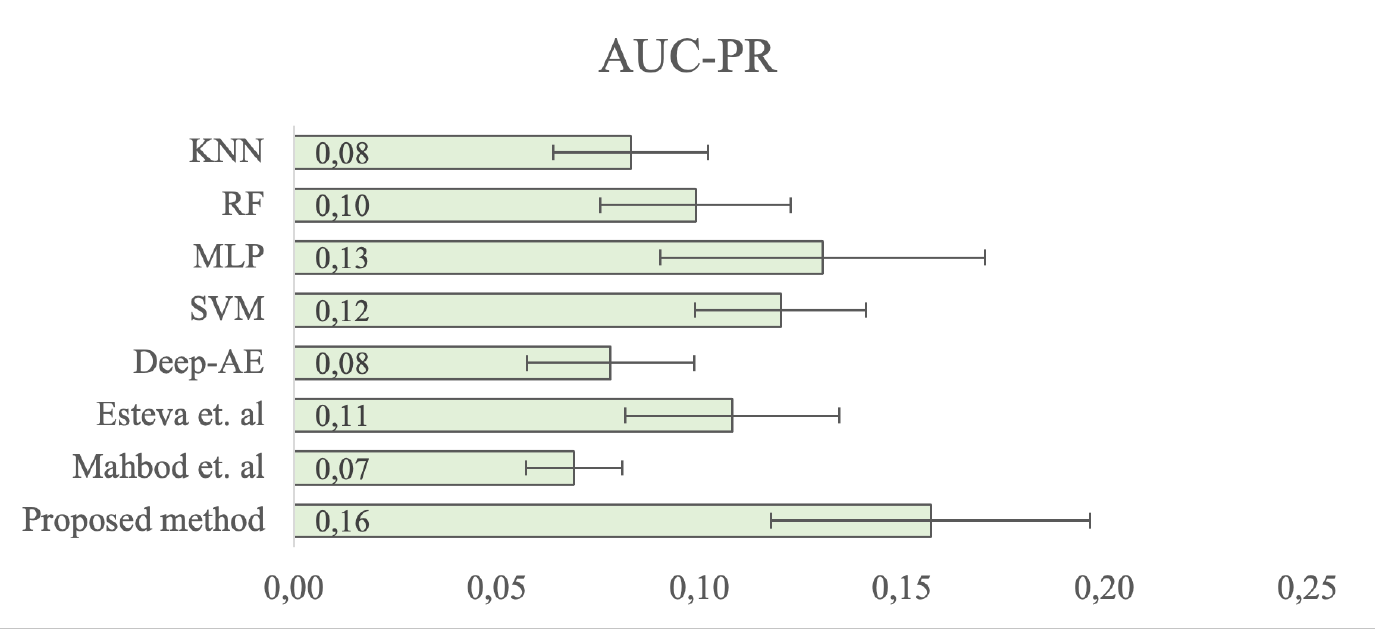}
%    \caption{Comparison of AUC-PR over ten independent runs}
    \label{comp2}
%\end{figure}
%\begin{figure}[H]
%    \centering
    \includegraphics[width=1\textwidth]{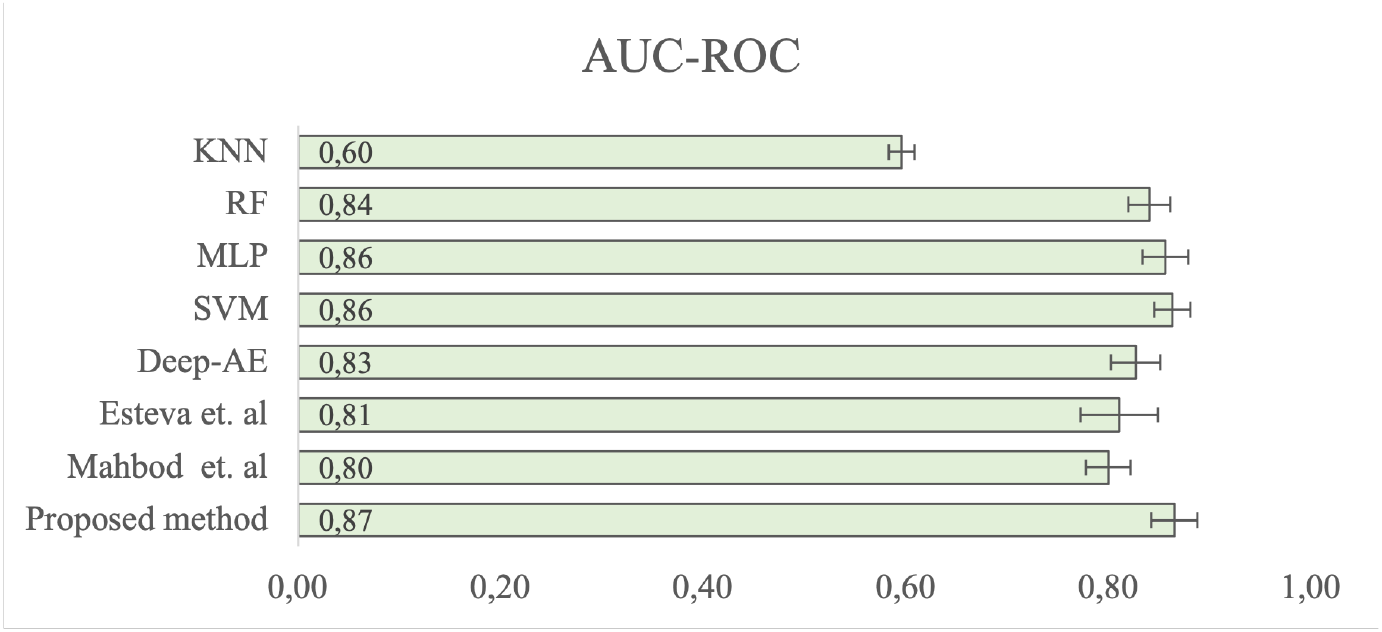}
    \caption{Classification accuracy of the proposed method and seven other methods measured by three evaluation metrics (F1-measure, AUC-PR, AUC-ROC). The scores are averages over $n=10$ independent repetitions. Error bars are 95 \% confidence intervals based on the $t$-distribution with $n-1=9$ degrees of freedom.}
    \label{comp1}
%    \label{comp3}
\end{figure}

Table \ref{comp} and Figure~\ref{comp1} show a comparison of the proposed technique with four non-deep learning classifiers (KNN, RF, MLP, SVM) and three selected deep learning based techniques; see Appendix B for the most important parameters of the benchmark methods.
In each of the benchmark methods except those by~\cite{19} and \cite{12}, the $32\times 32$ pixel RGB input images (altogether 3072 input features) along with the auxiliary information in the metadata (additional 3 input features) were used as the input.\footnote{The computational cost of training the SVM classifier prohibited the use of the higher-resolution images.}

\begin{figure}[tbp!]
    \centering
    \includegraphics[height=.39\textwidth]{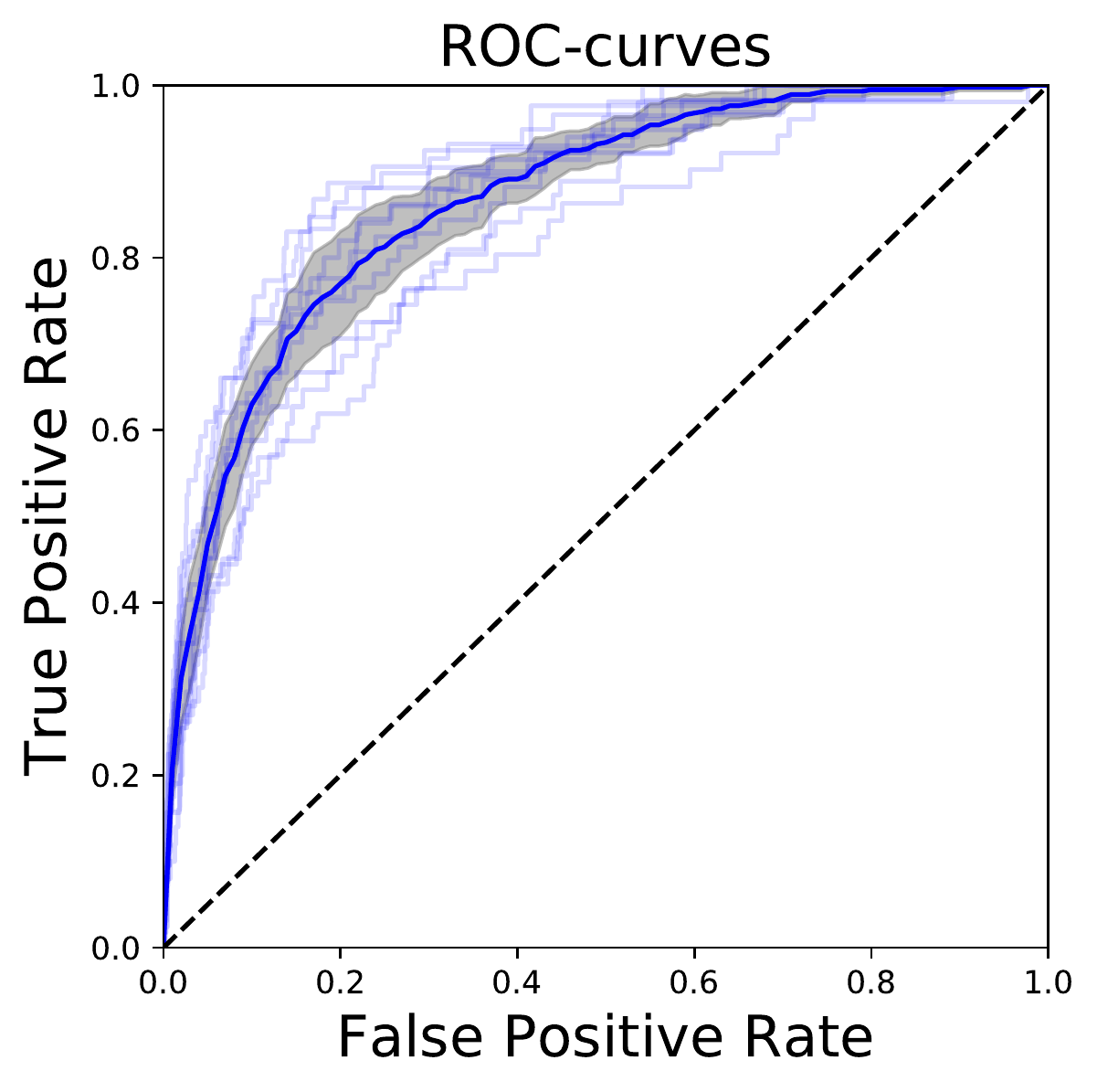}
    \includegraphics[height=.39\textwidth]{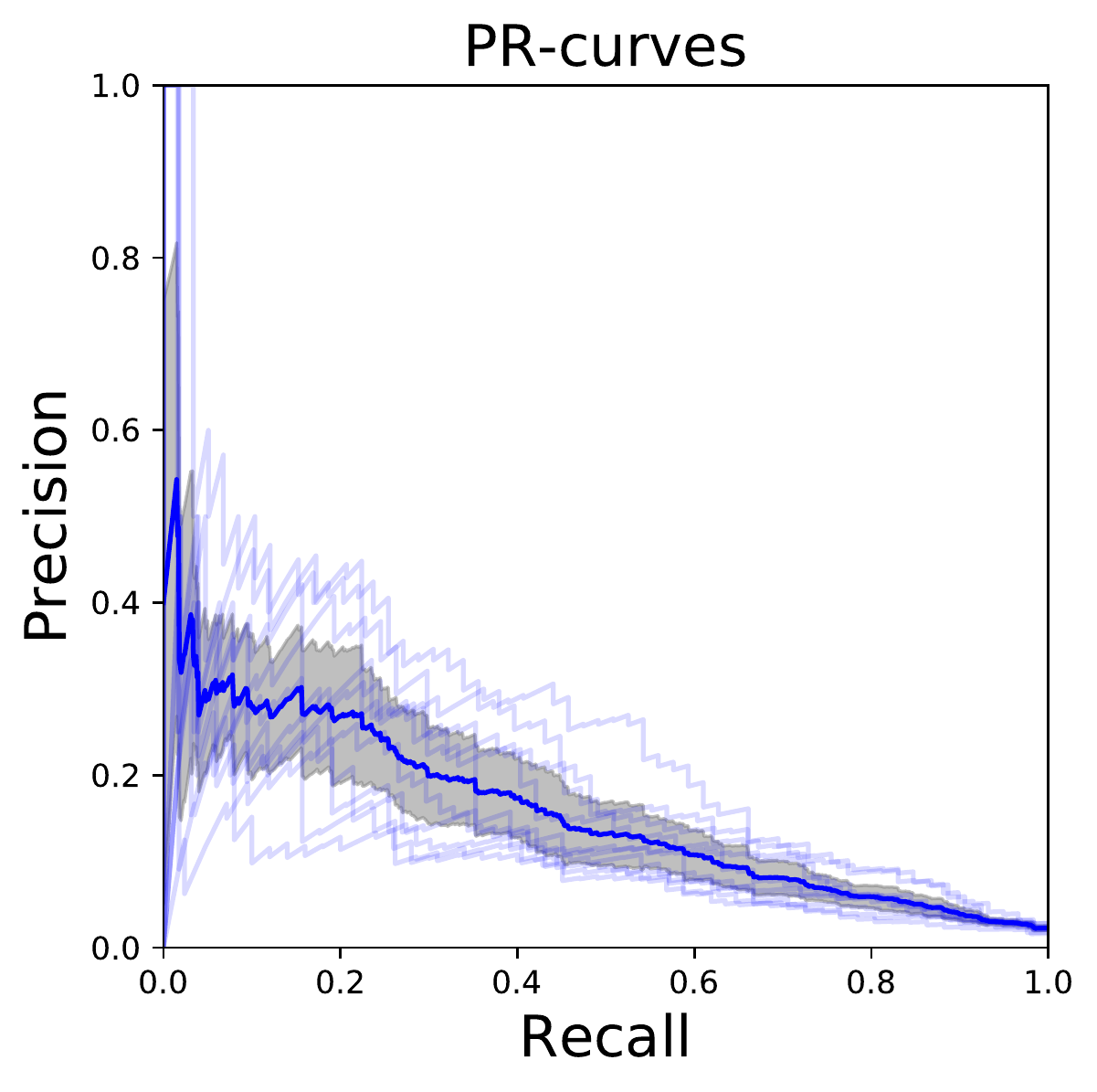}
    \caption{Left: AUC-ROC curves for the proposed method.  Right: AUC-PR curves for the proposed method. Both panels show the curves for ten independent runs (light blue curves), the average curve (bold blue line), and an interval showing the standard deviation of the curve (gray region).}
    \label{AUC-PR}
\end{figure}

\begin{figure}[tbp!]
    \centering
    \includegraphics[width=.49\textwidth]{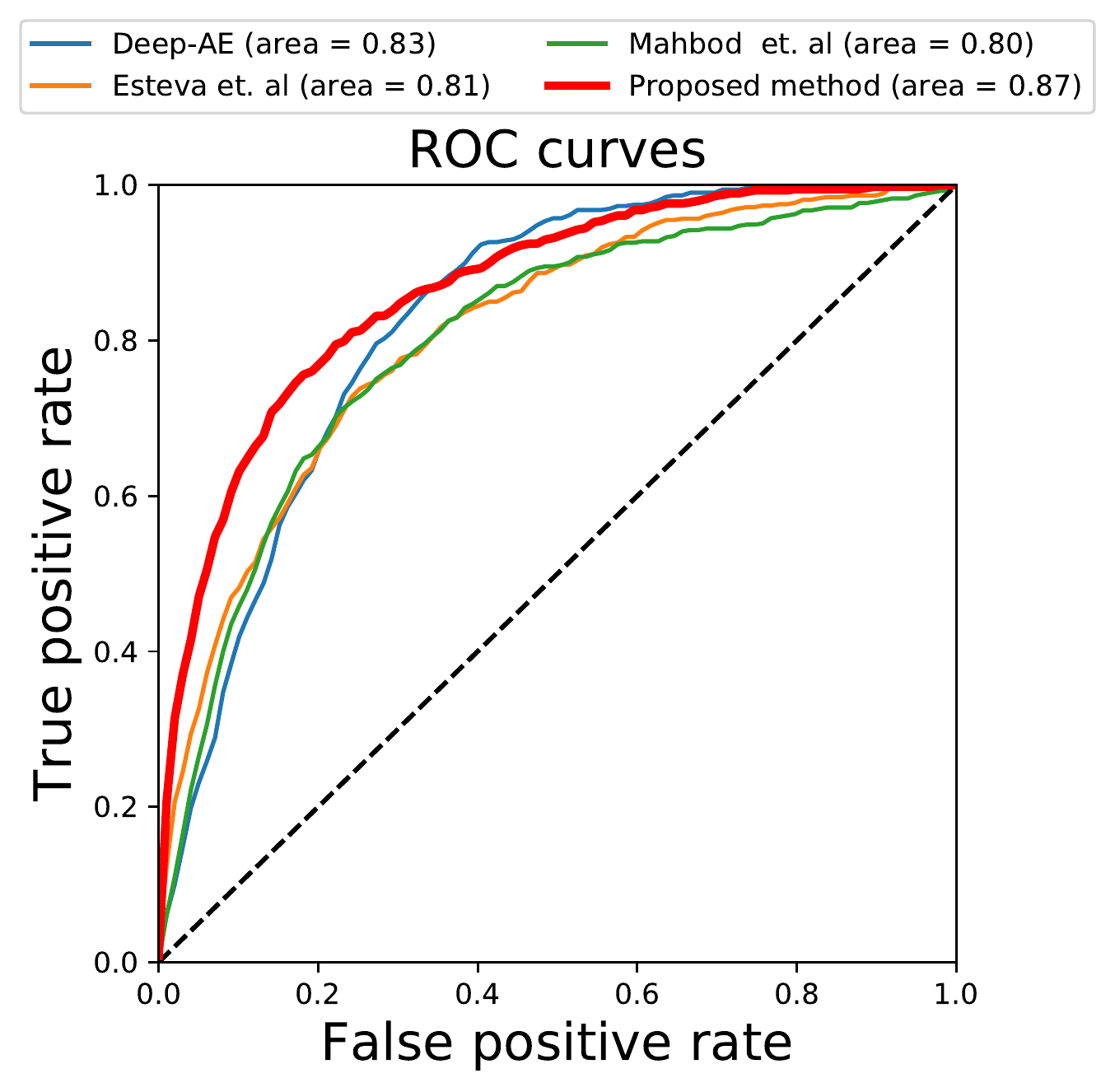}
    \includegraphics[width=.49\textwidth]{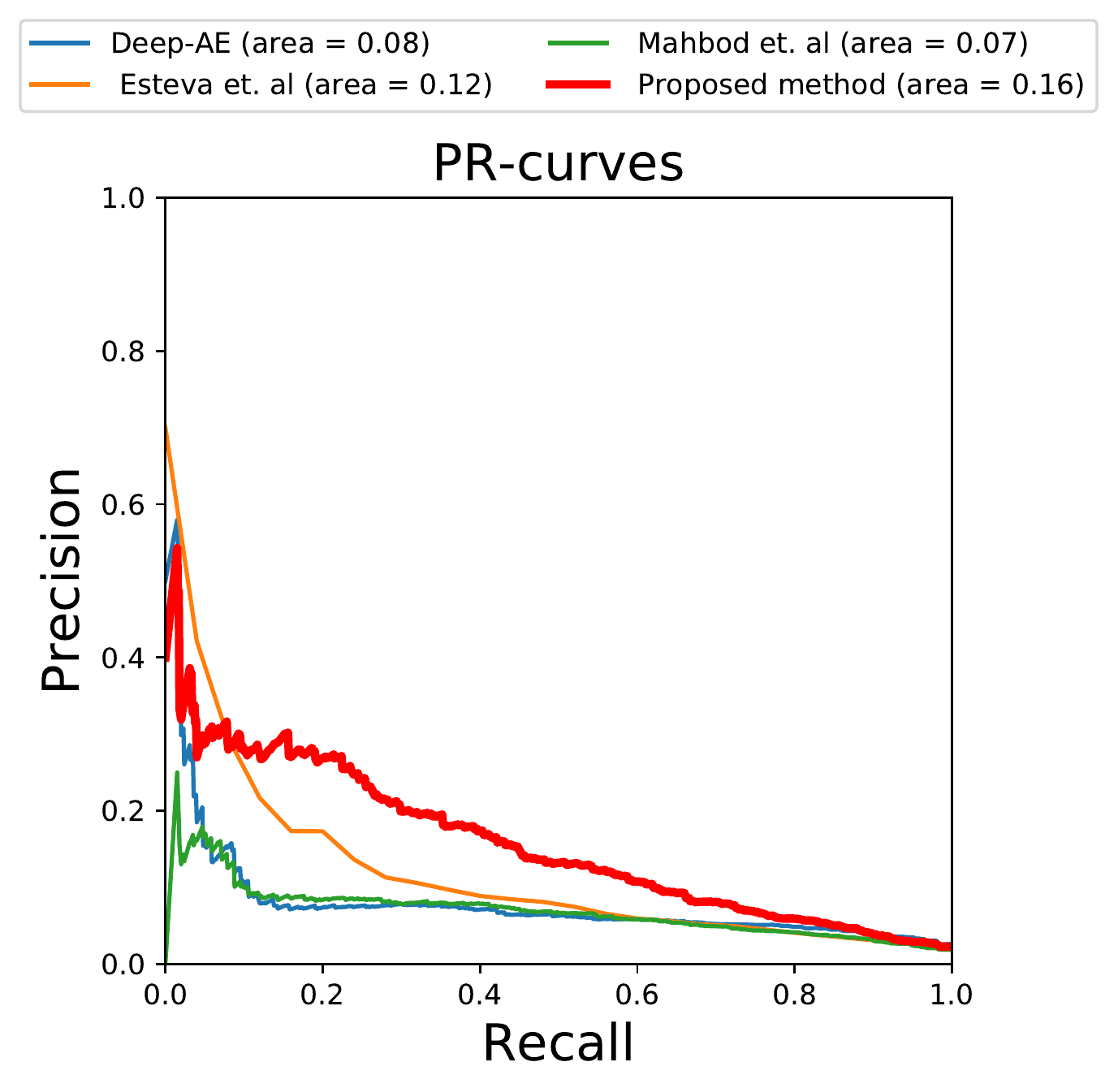}
    \caption{Left: Comparison of average ROC curves with three other deep learning methods. Right:  Average PR-curves.}
    \label{AUC-PR-STATE}
\end{figure}

\begin{figure}[tbp!]
    \centering
    \includegraphics[width=.49\textwidth]{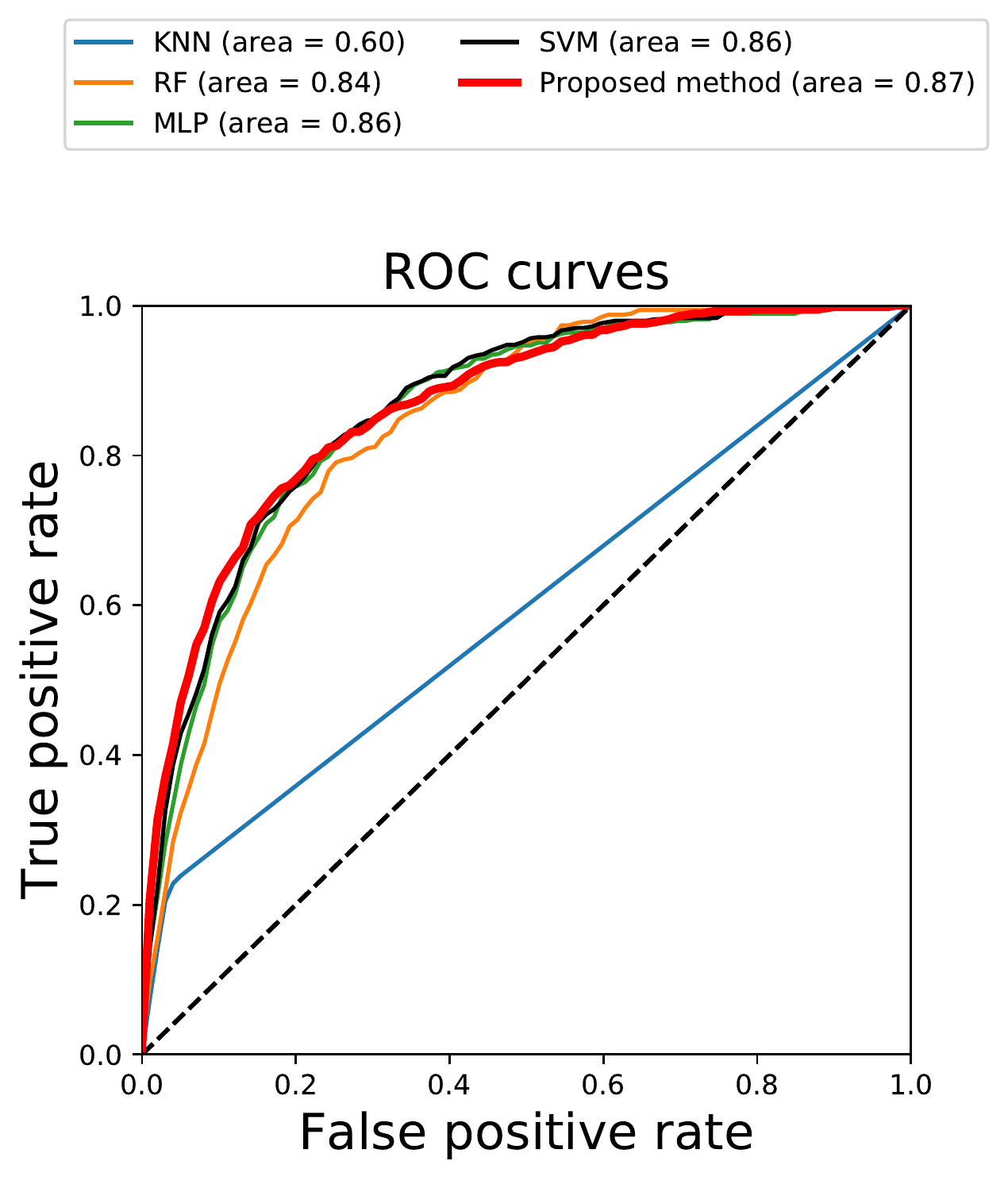}
    \includegraphics[width=.49\textwidth]{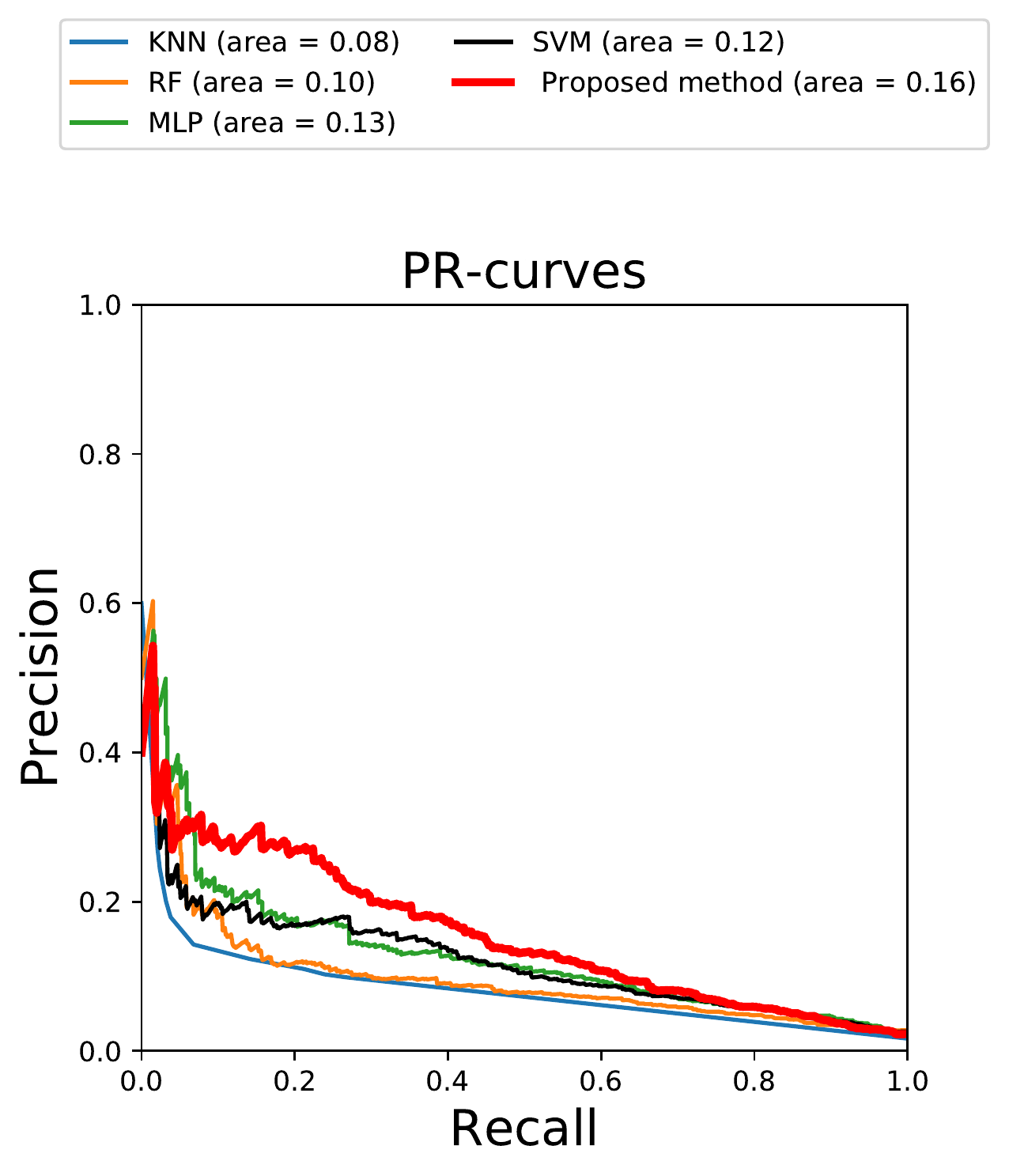}
    \caption{Left: Comparison of average ROC with non-deep learning methods. Right: Average PR-curves.}
    \label{AUC-PR-TECH}
\end{figure}

%Results show that the proposed technique is better in terms of all the evaluation measures. To check the stability of the all the techniques mentioned in Table \ref{comp}, the results are reported in terms of mean and standard deviation of error for ten independent runs. 
The proposed technique achieves average F1, AUC-PR, and AUC-PR values 0.23, 0.16, and 0.87 respectively, which are highest among all of the compared methods. %--\ref{comp3}. Figure 
However, the differences between the top performing methods are within statistical margin of error\footnote{Following~\cite{berrar-losano-2013}, we present comparisons in terms of confidence intervals instead of hypothesis tests (``We argue
that the conclusions from comparative classification studies should be based
primarily on effect size estimation with confidence intervals, and not on
significance tests and $p$-values.''). We calculate 95~\% confidence intervals based on the $t$-distribution with $n-1=9$ degrees of freedom by $\mu \pm 2.26216 \times {\sigma \over \sqrt{n}}$, where $\mu$ is the average score, $\sigma$ is the standard deviation of the score, and $n=10$ is the sample size (number of repetitions with independent random train-validation-test splits).}.
A more detailed visualization of the ROC and PR-curves is shown in Figs.~\ref{AUC-PR}--\ref{AUC-PR-TECH}.

\begin{table}[tbp!]
\caption{Comparison of the proposed method with seven other methods in terms of three evaluation metrics (F1-measure, AUC-PR, AUC-ROC). The table shows the average score over $n=10$ random repetitions $\pm$ 95 \% confidence intervals based on the $t$-distribution with $n-1=9$ degrees of freedom.} % title of Table
\centering
 \begin{tabular}{c c c c} 
 \toprule
 Method & F1-measure & AUC-PR & AUC-ROC \\ [0.5ex]\midrule
 KNN &  $0.13\pm0.02 $ &  $ 0.08\pm0.02  $ & $0.60\pm0.01 $  \\ 
 RF &  $0.05\pm0.02 $ &  $ 0.10\pm0.02  $ &  $ 0.84\pm0.02 $\\
 MLP &	 $ 0.13\pm0.03 $ &	 $ 0.13\pm0.04 $ & $ 0.86\pm0.02 $\\
 SVM &  $ 0.19\pm0.04 $ &  $ 0.12\pm0.02 $ &  $ 0.86\pm0.02 $ \\
 Deep-AE & $0.09\pm0.03 $ & $ 0.08\pm0.02 $ & $ 0.83\pm0.02 $ \\
 \cite{19} & $0.17\pm0.03$ &	$0.12\pm0.03$ & $0.81 \pm 0.04$ \\
 \cite{12} & $0.12\pm0.02$ &	$0.07\pm0.01$ & $0.80\pm0.02$ \\
 Proposed method & $\mathbf{0.23} \pm \mathbf{0.04}$ & $\mathbf{0.16} \pm \mathbf{0.04}$ & $\mathbf{0.87} \pm \mathbf{0.02}$ \\\bottomrule
\end{tabular}
\label{comp} % is used to refer this table in the text
\end{table}

\subsection{Performance gain from ensemble learning}

%Table \ref{baseline}. shows the performance of MLP, SVM, KNN, RF, and Deep Auto-Encoder (DAE) in terms of F1-measure, AUC-ROC, and AUC-PR for ten independent runs. Results show that SVM performs good among all the base-line classifiers.
%\begin{table}[H]
%\caption{Performance of base-line %classifiers} % title of Table
%\centering
% \begin{tabular}{c c c c} 
%\toprule
% Classifier & F1-measure & AUC-PR & %AUC-ROC \\ [0.5ex]\midrule
% KNN &  $0.13\pm0.03 $ &  $ 0.08\pm0.03  $ %& $0.60\pm0.02 $  \\ 
% RF &  $0.05\pm0.03 $ &  $ 0.10\pm0.03  $ %&  $ 0.84\pm0.03 $\\
% Deep-AE & $0.09\pm0.04 $ & $ 0.08\pm0.03 %$ & $ 0.83\pm0.03 $ \\
% MLP &	 $ 0.13\pm0.05 $ &	 $ 0.13\pm0.05 %$ & $ 0.86\pm0.03 $\\
% SVM &  $ 0.19\pm0.05 $ &  $ 0.12\pm0.03 $ %&  $ 0.86\pm0.03 $ \\\bottomrule
%\end{tabular}
%\label{baseline} % is used to refer this %table in the text
%\end{table}
%\subsection{Performance of base-learners and meta-classifier in terms of  F1-measure, AUC-PR, and AUC-ROC}

The proposed technique in comprised of two steps; in first step base-learners are trained and in second step meta-learner is trained on the top of base-learners. %Therefore, it is important to show the performance comparison of base-learners and meta classifier. 
Table \ref{base_meta} shows the performance of each of the base-learners individually, which can be compared with the performance of the resulting SVM meta-classifier that combines the base-learners outputs as the ensemble classification. Out of six base-learners four are trained from scratch on the ISIC 2020 dataset while the remaining two are pre-trained on skin cancer images that are not part of ISIC 2020 dataset. The performance comparison  shows that even though the accuracies of each of the base-learners individually are quite low, the meta-classifier performs markedly better. This suggests that the base-learners succeed in providing a diverse set of inputs to the meta-learner, thus significantly improving overall performance of the ensemble over any of the base-learners. 
%Table \ref{base_meta} shows that each base-learner is different in terms of performance and this difference in opinion diversifies the feature space provided to meta-classifier. Therefore, meta classifier shows significantly better performance. 

\begin{table}[tbp!]
\caption{Performance of the base-learners and the full ensemble (the SVM meta-classifier), showing the significantly better performance by the ensemble compared to the individual base-learners.} % title of Table
\centering
 \begin{tabular}{c c c c} 
\toprule
Learner  & F1-Measure & AUC-PR & AUC-ROC \\ [0.5ex]\midrule
$\mathrm{CNN}_{32\times32}$ & $0.09\pm0.03 $ & $0.10\pm0.03$  & $0.82\pm0.04$ \\
$\mathrm{CNN}_{64\times64}$ & $0.12\pm0.02$ & $0.11\pm0.03$ & $0.83\pm0.02$ \\
$\mathrm{CNN}_{128\times128}$ & $0.13\pm0.02$ & $0.11\pm0.03$ & $0.85\pm0.02$ \\
$\mathrm{CNN}_{256\times_256}$ & $0.11\pm0.01$ & $0.09\pm0.02$ & $0.84\pm0.04$ \\
pre-trained $\mathrm{CNN}_{32\times32}$ & $0.08\pm0.01$ & $0.04\pm0.01$ & $0.72\pm0.03$ \\
pre-trained $\mathrm{CNN}_{64\times64}$ & $0.07\pm0.01$ & $0.03\pm0.01$ & $0.71\pm0.03$ \\
ensemble (SVM meta-classifier) & \textbf{0.23}$\pm$\textbf{0.04} & \textbf{0.16}$\pm$\textbf{0.04} & \textbf{0.87}$\pm$\textbf{0.02} \\\bottomrule
\end{tabular}
\label{base_meta} % is used to refer this table in the text
\end{table}

\subsection{Significance of transfer learning and meta-data} % in the proposed technique}

%In the proposed technique six different CNNs are used as base-learners. 
%During training four out of six base-learners are trained from scratch. Whereas two of base-learners are pre-trained on images collected from ISIC archive and have equal number of malignant and benign images. 
%The purpose of combining pre-trained CNNs trained on a balanced dataset with CNNs trained fully on the ISIC 2020 data is to provide a diverse set of predictions to the meta-learner. We also include the metadata available in the ISIC 2020 data as an additional source of diversity to the meta-learner. 

To evaluate the impact of the using pre-trained models and that of the metadata on the classification accuracy in the proposed method, we also evaluated the performance with either one of these components disabled. Table \ref{NTL} shows the performance comparison of the proposed technique to a version where the pre-trained CNNs are disabled, and one where the metadata is not included as auxiliary data for the meta-learner. As seen in the table, excluding the pre-trained CNNs doesn't significantly affect the performance. The exclusion of the metadata led to somewhat inferior performance, but here too, the differences are relatively minor and with within statistical margin of error. Further research with larger datasets and richer metadata is needed to confirm the benefits.

\begin{table}[tbp!]
\caption{Performance comparison of the proposed technique without pre-trained base-learners and meta-data} % title of Table
\centering
 \begin{tabular}{c c c c}
 \toprule
Method & F1-measure & AUC-PR & AUC-ROC \\ [0.5ex]\midrule
 without pre-trained base-learners &	$\mathbf{0.23}\pm\mathbf{0.04}$	& $0.14\pm0.03$ &	$\mathbf{0.87}\pm\mathbf{0.02}$  \\ 
 without meta-data & $0.17\pm0.03$ & $0.13\pm0.04$ & $0.86\pm0.02$
 \\ \midrule
 with both (= proposed method) & $\mathbf{0.23} \pm \mathbf{0.04}$ & $\mathbf{0.16} \pm \mathbf{0.04}$ & $\mathbf{0.87} \pm \mathbf{0.02}$ \\\bottomrule
\end{tabular}
\label{NTL} % is used to refer this table in the text
\end{table}

\section{Conclusions}

We proposed an ensemble-based deep learning approach for skin cancer detection based on dermoscopic images.  Our method uses an ensemble of convolutional neural networks (CNNs) trained of input images of different sizes along with metadata. 
We present results on the ISIC 2020 dataset which is contains 33126 dermoscopic images from 2056 patients. The dataset is highly imbalanced with less than 2~\% of malignant samples. The impact of ensemble learning was found to be significant, while the impact of transfer learning and the use of auxiliary information in the form of metadata associated with the input images appeared to be minor. The proposed method compared favourably against other machine learning based techniques including three deep learning based techniques, making it a promising approach for skin cancer detection especially on imbalanced datasets. Our research expands the evidence suggesting that deep learning techniques offer useful tools in dermatology and other medical applications.

%In this proposed technique two base-learners are pre-trained on balanced skin cancer data collected from ISIC archive (these images are not part of ISIC 2020 data set), while rest of base learners are trained from scratch on ISIC 2020 data set (which is imbalanced). Most of the times, data augmentation or down sampling is used for data balancing. In case of data augmentation artificial samples are generated which have limited information and not have as much precise information as real data have. Whereas down sampling leads to loss of information that original data has. Unlike data augmentation and down sampling techniques, there is no loss of information because base-learners trained from scratch on ISIC 2020 data extract useful information and this knowledge is further boosts by incorporating the information that is extracted from the pre-trained base-learners. Therefore, all the base-learners generate different hypothesis spaces and thus help the meta-learner (SVM) to make a robust decision for skin cancer classification. 

\section*{Acknowledgements}

This work is supported by the Academy of Finland (Projects TensorML \#311277, and the FCAI Flagship)

\bibliography{mybibfile}

\section*{Appendix A: Neural Network Architectures and Hyperparameters of the Proposed Method}
\setcounter{table}{0}
\renewcommand{\thetable}{A\arabic{table}}
\begin{table}[H]
 \scriptsize
\caption{Parameter setting of $\mathrm{CNN}_{32\times32}$} % title of Table
\centering
 \begin{tabular}{c c c c c c}
 \toprule
Layer & Layer &	Kernel size & Feature maps and neurons & Stride	& Activation function \\ \midrule
0 &	Input &	--- & $3\times32\times32$ &	--- &	Relu  \\ 
1 &	Convolutional &	$2\times2$ & $64\times32\times32$ & [1 ,1] &	Relu  \\ 
2 &	Max pooling & --- &  $64\times16\times16$ & [2 ,2] & --- \\
3 &	Dropout &	--- & $64\times16\times16$ & ---	& --- \\
4 &	Convolutional &	$3\times3$	& $40\times16\times16$ & [1 ,1] &	Relu \\
5 & Max Pooling & --- &  $40\times8\times8$ & [2 ,2] & --- \\
6 &	Convolutional &	$3\times3$	& $30\times8\times8$ & [1 ,1] &	Relu\\
7 &	Max pooling &	--- &  $30\times4\times4$ & [2 ,2] & ---\\
8 & Convolutional &	$3\times3$ & $25\times4\times4$ & [1 ,1] &	Relu \\
9 &	Max Pooling &	--- &  $25\times2\times2$ & [2 ,2] & --- \\
10 & Fully connected &	--- &	512 & --- &	Sigmoid \\
11 & Dropout & --- & 512 & --- & --- \\
12	& Fully connected &	---	& 1& --- &	Sigmoid \\\bottomrule
\end{tabular}
\label{CNN1} % is used to refer this table in the text
\end{table}

\begin{table}[H]
 \scriptsize
\caption{Parameter setting of $\mathrm{CNN}_{64\times64}$} % title of Table
\centering
 \begin{tabular}{c c c c c c}
 \toprule
Layer & Layer &	Kernel size & Feature maps and neurons & Stride	& Activation function \\ [0.5ex] \midrule
0 &	Input &	--- & $3\times64\times64$ &	--- &	---  \\ 
1 &	Convolutional &	$2\times2$ & $128\times64\times64$ & [1 ,1] &	Relu  \\

2 &	Max pooling & --- &  $128\times32\times32$ & [2 ,2] & --- \\
3 &	Dropout &	--- & $128\times32\times32$ & ---	& --- \\
4 &	Convolutional &	$3\times3$	& $80\times32\times32$ & [1 ,1] &	Relu \\
5 & Max Pooling & --- &  $80\times16\times16$ & [2 ,2] & --- \\
6 &	Convolutional &	$3\times3$	& $60\times16\times16$ & [1 ,1] &	Relu\\
7 &	Max pooling &	--- &  $60\times8\times8$ & [2 ,2] & ---\\
8 & Convolutional &	$3\times3$ & $50\times8\times8$ & [1 ,1] &	Relu \\
9 &	Max Pooling &	--- &  $50\times4\times4$ & [2 ,2] & --- \\
10 & Fully connected &	--- &	512 & --- &	Sigmoid \\
11 & Dropout & --- & 512 & --- & --- \\
12	& Fully connected &	---	& 1& --- &	Sigmoid \\\bottomrule
\end{tabular}
\label{CNN2} % is used to refer this table in the text
\end{table}

\begin{table}[H]
\scriptsize
\caption{Parameter setting of $\mathrm{CNN}_{128\times128}$} % title of Table
\centering
 \begin{tabular}{c c c c c c} 
 \toprule
Layer & Type of layer &	Kernel size & No of feature maps and neurons & Stride &	Activation function \\ [0.5ex] \midrule
0 &	Input &	--- & $3\times128\times128$ &	--- &	---  \\ 
1 &	Convolutional &	$1\times1$ & $192\times128\times128$ & [1 ,1] &	Relu  \\
2 &	Max pooling & --- &  $192\times64\times64$ & [2 ,2] & --- \\
3 &	Dropout &	--- & $192\times64\times64$ & ---	& --- \\
4 &	Convolutional &	$3\times3$	& $120\times64\times64$ & [1 ,1] &	Relu \\
5 & Max Pooling & --- &  $120\times32\times32$ & [2 ,2] & --- \\
6 &	Convolutional &	$3\times3$	& $90\times32\times32$ & [1 ,1] &	Relu\\
7 &	Max pooling &	--- &  $90\times16\times16$ & [2 ,2] & ---\\
8 & Convolutional &	$3\times3$ & $75\times16\times16$ & [1 ,1] &	Relu \\
9 &	Max Pooling &	--- &  $75\times8\times8$ & [2 ,2] & --- \\
10 & Fully connected &	--- &	512 & --- &	Sigmoid \\
11 & Dropout & --- & 512 & --- & --- \\
12	& Fully connected &	---	& 1& --- &	Sigmoid \\\bottomrule
\end{tabular}
\label{CNN3} % is used to refer this table in the text
\end{table}

\begin{table}[H]
\scriptsize
\caption{Parameter setting of $\mathrm{CNN}_{256\times256}$} % title of Table
\centering
 \begin{tabular}{c c c c c c} 
 \toprule
Layer & Type of layer &	Kernel size & No of feature maps and neurons &	Stride & Activation function \\ [0.5ex] \midrule
 0 &	Input &	--- & $3\times256\times256$ &	--- &	---  \\ 
1 &	Convolutional &	$2\times2$ & $256\times256\times256$ & [1 ,1] &	Relu  \\
2 &	Max pooling & --- &  $256\times128\times128$ & [2 ,2] & --- \\
3 &	Dropout &	--- & $256\times128\times128$ & ---	& --- \\
4 &	Convolutional &	$3\times3$	& $160\times128\times128$ & [1 ,1] & Relu\\
5 & Max Pooling & --- &  $160\times64\times64$ & [2 ,2] & --- \\
6 &	Convolutional &	$3\times3$	& $120\times64\times64$ & [1 ,1] &	Relu\\
7 &	Max pooling &	--- &  $120\times32\times32$ & [2 ,2] & ---\\
8 & Convolutional &	$3\times3$ & $100\times32\times32$ & [1 ,1] &	Relu \\
9 &	Max Pooling &	--- &  $100\times16\times16$ & [2 ,2] & --- \\
10 & Fully connected &	--- &	512 & --- &	Sigmoid \\
11 & Dropout & --- & 512 & --- & --- \\
12	& Fully connected &	---	& 100 & --- &	Sigmoid \\
13	& Fully connected &	---	& 1 & --- &	Sigmoid \\\bottomrule
\end{tabular}
\label{CNN4} % is used to refer this table in the text
\end{table}

\begin{table}[H]
\scriptsize
\caption{Parameter setting of pre-trained $\mathrm{CNN}_{32\times32}$} % title of Table
\centering
 \begin{tabular}{c c c c c c} 
  \toprule
Layer & Type of layer &	Kernel size & No of feature maps and neurons &	Stride & Activation function \\ [0.5ex]\midrule
0 &	Input &	--- & $3\times32\times32$ &	--- &	Relu  \\ 
1 &	Convolutional &	$2\times2$ & $200\times32\times32$ & [1 ,1] &	Relu  \\
2 &	Max pooling & --- &  $200\times16\times16$ & [2 ,2] & --- \\
3 &	Dropout &	--- & $200\times16\times16$ & ---	& --- \\
4 &	Convolutional &	$3\times3$	& $80\times16\times16$ & [1 ,1] &	Relu \\
5 & Max Pooling & --- &  $80\times8\times8$ & [2 ,2] & --- \\
6 &	Convolutional &	$3\times3$	& $60\times8\times8$ & [1 ,1] &	Relu\\
7 &	Max pooling &	--- &  $60\times4\times4$ & [2 ,2] & ---\\
8 & Convolutional &	$3\times3$ & $50\times4\times4$ & [1 ,1] &	Relu \\
9 &	Max Pooling &	--- &  $50\times2\times2$ & [2 ,2] & --- \\
10 & Fully connected &	--- &	1024 & --- &	Sigmoid \\
11 & Dropout & --- & 1024 & --- & --- \\
12	& Fully connected &	---	& 1& --- &	Sigmoid \\\bottomrule 
\end{tabular}
\label{CNN5} % is used to refer this table in the text
\end{table}

\begin{table}[H]
\scriptsize
\caption{Parameter setting of pre-trained $\mathrm{CNN}_{64\times64}$} % title of Table
\centering
 \begin{tabular}{c c c c c c} 
 \toprule
Layer & Type of layer &	Kernel size & No of feature maps and neurons &	Stride & Activation function \\ [0.5ex]\midrule
0 &	Input &	--- & $3\times64\times64$ &	--- &	Relu  \\ 
1 &	Convolutional &	$2\times2$ & $128\times64\times64$ & [1 ,1] &	Relu  \\
2 &	Max pooling & --- &  $128\times32\times32$ & [2 ,2] & --- \\
3 &	Dropout &	--- & $128\times32\times32$ & ---	& --- \\
4 &	Convolutional &	$2\times2$	& $80\times32\times32$ & [1 ,1] &	Relu \\
5 & Max Pooling & --- &  $80\times16\times16$ & [2 ,2] & --- \\
6 &	Convolutional &	$2\times2$	& $100\times16\times16$ & [1 ,1] &	Relu\\
7 &	Max pooling &	--- &  $100\times8\times8$ & [2 ,2] & ---\\
8 & Convolutional &	$2\times2$ & $70\times8\times8$ & [1 ,1] &	Relu \\
9 &	Max Pooling &	--- &  $70\times4\times4$ & [2 ,2] & --- \\
10 & Fully connected &	--- &	700 & --- &	Sigmoid \\
11 & Dropout & --- & 700 & --- & --- \\
12	& Fully connected &	---	& 1& --- &	Sigmoid \\\bottomrule
\end{tabular}
\label{CNN6} % is used to refer this table in the text
\end{table}

\begin{table}[H]
\caption{Parameters of the meta-learner (Support Vector Machine)} % title of Table
\centering
 \begin{tabular}{c c c c} 
 \toprule
 Kernel &	Degree & $C$ &	$\gamma$ \\ \midrule
 rbf & 2	& 0.02 & 0.0009 \\ 
 \bottomrule
\end{tabular}
\label{svm-para} % is used to refer this table in the text
\end{table}

\section*{Appendix B: Hyperparameters of the Compared Methods}
\setcounter{table}{0}

We used the following hyperparameters for the compared methods, which were selected manually by adjusting them until no further improvement on the validation data performance (F1-measure) was observed. For the KNN classifier, we use $k=4$. For the random forest (RF), maximum depth 30 and $n=100$ trees are used. For the multilayer perceptron (MLP), we use one hidden layer with 120 neurons, and minibatch size 250. For the support vector machine (SVM), a polynomial kernel of degree 3, and constants $C=0.07$ and $\gamma=0.0009$ are used. In the deep autoencoder (deep-AE), we use an encoder with two layers having 2000 and 1000 neurons, respectively, and a symmetric decoder, and train the model for 100 epochs with minibatch size 15.

\end{document}